\documentclass[letterpaper, 10 pt, conference]{ieeeconf}  

\IEEEoverridecommandlockouts                              

\usepackage{glossaries}
\usepackage{graphicx}
\usepackage{svg}
\usepackage{multirow}
\usepackage{fontawesome5}
\usepackage{xcolor}

\makeglossaries

\newglossaryentry{NeRF}{
  name=Neural Radiance Fields (NeRFs),
  description={A document preparation system}
}

\title{\LARGE \bf
AgriNeRF: Neural Radiance Fields for Agriculture in Challenging Lighting Conditions
}

\author{Samarth Chopra$^\dagger$, Fernando Cladera$^\ddagger$, Varun Murali$^\ddagger$, and Vijay Kumar$^\ddagger$
\thanks{
We gratefully acknowledge the support of the IoT4Ag Engineering Research Center funded by the National Science Foundation (NSF) under NSF Cooperative Agreement Number EEC-1941529, 
NIFA grant 2022-67021-36856, 
NSF grant CCR-2112665, 
and ARL DCIST CRA W911NF-17-2-0181,
and NVIDIA.
}
\thanks{
$^\dagger$University of Pittsburgh,
    {\tt\small sac345@pitt.edu}}%
\thanks{
$^\ddagger$GRASP Laboratory, University of Pennsylvania, Philadelphia, PA, 19104, USA 
    {\tt\small \{fclad, mvarun, kumar\}@seas.upenn.edu}}%
}

\begin{document}

\maketitle
\thispagestyle{empty}
\pagestyle{empty}

\begin{abstract}


\gls{NeRF} have shown significant promise in 3D scene reconstruction and novel view synthesis. In agricultural settings, NeRFs can serve as digital twins, providing critical information about fruit detection for yield estimation and other important metrics for farmers. However, traditional NeRFs are not robust to challenging lighting conditions, such as low-light, extreme bright light and varying lighting. 
To address these issues, this work leverages three different sensors: an RGB camera, an event camera and a thermal camera.
Our RGB scene reconstruction shows an improvement in PSNR and SSIM by +2.06 dB and +8.3\% respectively. Our cross-spectral scene reconstruction enhances downstream fruit detection by +43.0\% in mAP50 and +61.1\% increase in mAP50-95.
The integration of additional sensors leads to a more robust and informative NeRF. We demonstrate that our multi-modal system yields high quality photo-realistic reconstructions under various tree canopy covers and at different times of the day. This work results in the development of a resilient NeRF, capable of performing well in visibly degraded scenarios, as well as a learnt cross-spectral representation, that is used for automated fruit detection.

\end{abstract}

\section{INTRODUCTION}
In the context of precision agriculture, a digital twin of a farm or forest can provide high-resolution real-time analytics to farmers and foresters. Digital twins enable crop-monitoring from camera-equipped UAVs and UGVs, for optimization of agricultural practices like irrigation, fertilization, and crop harvesting~\cite{tagarakis2024digital}. Digital twins for automated fruit detection can enable yield estimation and mapping~\cite{bargoti2017deep}. This information is highly valuable as it can promote economic utilization of resources, resulting in higher returns per unit area and time [2].

\gls{NeRF} \cite{mildenhall2021nerf} have demonstrated incredible fidelity in capturing 3D scenes and novel view synthesis. Their ability to represent a full 3D scene using only a sparse set of 2D views allows for the creation of scalable digital twins. This is especially useful for farms and forests, which are hundreds of acres in size. Recent works \cite{hu2024high} \cite{fruitnerf2024} have shown the potential of NeRFs for phenotyping in agricultural environments to measure the morphological traits of plants, as well as for applications like automated fruit detection. In addition to reconstructing coarse details, NeRFs also show great performance in capturing finer details in scenes. This make them particularly suitable for fruit detection, where fruits can be only centimetres in size \cite{mildenhall2021nerf}. 

\begin{figure}[t!]
    \centering
\includegraphics[width=0.48\textwidth]{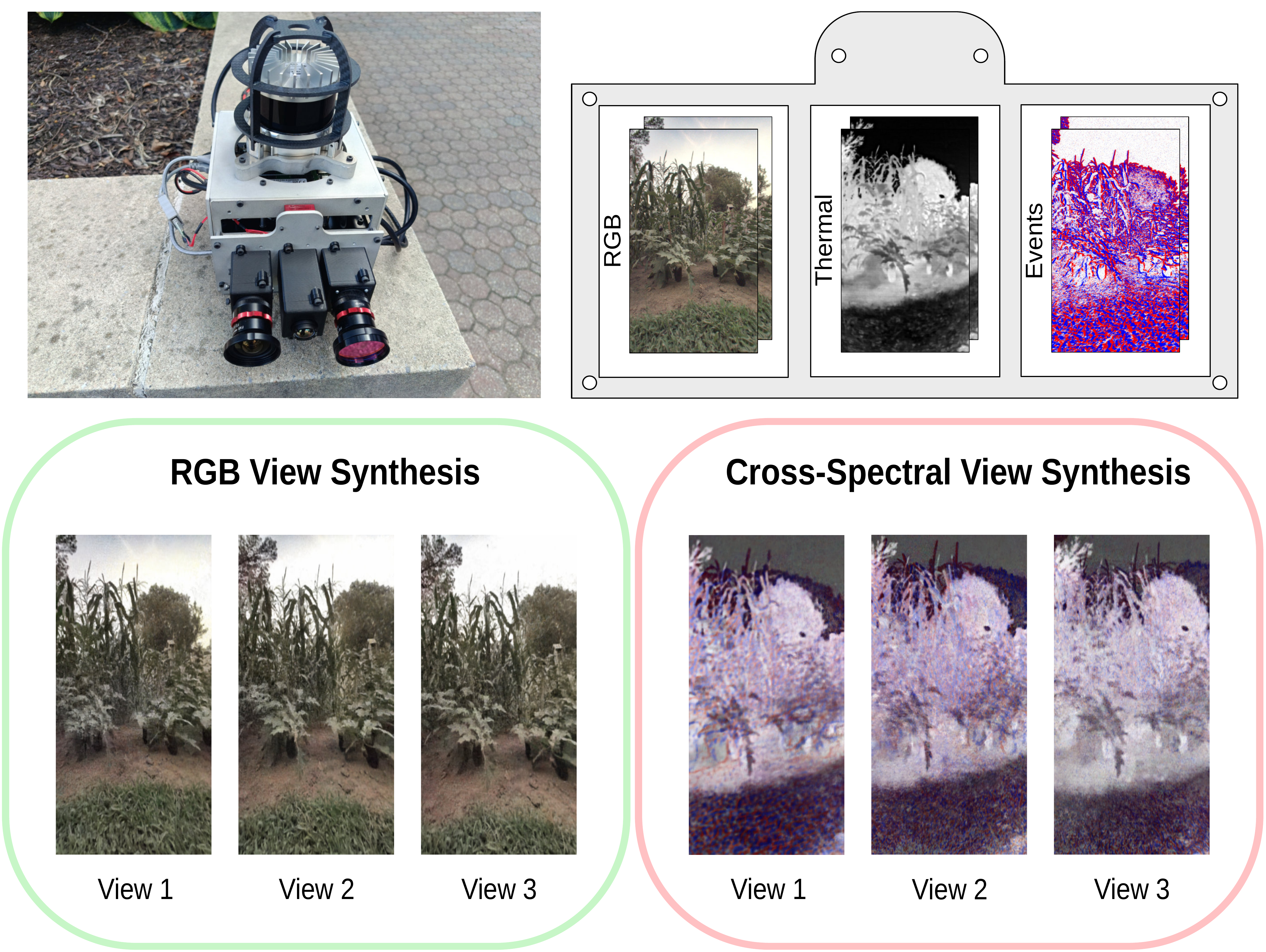}
\caption{
\textit{AgriNeRF} combines information from RGB, thermal and event cameras for resilient NeRF estimation in degraded lighting environments.
\underline{\textbf{Top}}: Sensor suite and input data. \underline{Left}: Multi-modal sensor suite equipped with three different sensors. \underline{Right}: RGB, thermal and event camera frames generated by sensor suite. These frames correspond to the physical camera location on the sensor suite.
\textbf{\underline{Bottom}}: Frames from the three cameras are used to generate RGB and Cross-Spectral NeRFs.
}
\vspace{-.7cm}
\label{fig:figure1}
\end{figure}

Traditional NeRFs however, struggle to render sharp reconstructions when affected by challenging lighting conditions, such as low-light, extreme bright light and varying lighting. 
These problems are extremely prevalent in robotics, particularly in farms and forests, where a robot may be driving or flying under a canopy with low-light or in other degraded illumination conditions. This is a major concern, as these systems need to be resilient to such factors for deployment in real-life applications.

Recently, event cameras have emerged as novel sensors that capture information similarly to the human eye. These sensors are robust to motion blur, due to their high temporal resolution (on the order of $\mu$s), and high pixel bandwidth (on the order of kHz)~\cite{gallego2020event}. Due to their inherent properties, event cameras have seen impressive application in NeRFs for image deblurring \cite{Cannici_2024_CVPR} \cite{klenk2023nerf}. These cameras also have very high dynamic range (140 dB vs. 60 dB) which means that they can adapt to bright and dark stimuli~\cite{gallego2020event}. Klenk et al.~\cite{klenk2023nerf} have shown some success in recovering extremely dark scenes with event-aided NeRFs.

Despite their high dynamic range, event cameras still require visible light. In scenarios with no visible light, these sensors are of little use. Thermal cameras are sensors that operate using infrared light, and can thus capture objects with a temperature above absolute zero. These sensors have seen use in military and civilian sectors, particularly for scenarios with total darkness \cite{gade2014thermal}. Some works have used thermal cameras for recovering NeRFs, and demonstrated substantial results in low-light situations~\cite{ye2024thermal, hassan2024thermonerf}. 

In this work, depicted in Fig. \ref{fig:figure1}, we propose \textit{AgriNeRF}, a novel NeRF framework employing information from three different cameras: RGB, thermal and event cameras.
The purpose of this multi-modal approach is two-fold. Firstly, we attempt to improve the RGB reconstruction of the scene using information from all three sensors. To do this, we introduce a regularization loss term, to maximize feature correlation between RGB, event and thermal frames. Secondly, we jointly train a cross-spectral reconstruction of the scene, which incorporates features from both the visible and infrared bands of the electromagnetic spectrum. This representation lends itself well to robust fruit detection. We validate our work on self-collected real-world novel datasets, using a self-built sensor suite, incorporating RGB, thermal, event cameras. 

In this paper, we present \textbf{two distinct contributions} to the fields of agricultural robotics and Neural Radiance Fields.

\begin{itemize}

 \item For the first time, we show how RGB, event and thermal cameras can be leveraged to construct Neural Radiance Fields, specifically for fruit detection. We show marked improvement in scene reconstruction by +2.06 dB in PSNR and a +8.3\% increase in SSIM relative to prior works.
 \item We  created two novel real-world multi-modal datasets, collected from apple and peach orchards, as well as gardens and parks, incorporating RGB, event, thermal, LiDAR and IMU data.
\end{itemize} 

We note that our cross-spectral reconstructions shows a +44.8\% increase in mAP50, a +55.7\% increase in mAP50-95, a +36.5\% increase in recall, and a +37.9\% increase in F1-score.

\section{Related Work}

\subsection{Neural Radiance Fields (NeRFs)}

\gls{NeRF}~\cite{mildenhall2021nerf} have emerged as a state-of-the-art method for 3D scene reconstruction and novel view synthesis. Given a sparse set of input views, NeRFs can represent a scene by querying 5D coordinates ($x, y, z, \theta, \phi$) along camera rays. Classical volume rendering techniques are then used to create output densities and view-dependent colors for a scene.

In recent times, there have been many architectural innovations to make NeRFs more robust to visible degraded scenes and motion blur \cite{mildenhall2022nerf} \cite{ma2022deblur} \cite{peng2022pdrf} \cite{lee2023dp}. Instead of using tonemapped low dynamic range images as input, RawNeRF \cite{mildenhall2022nerf} uses raw images to handles scenes with high dynamic range or those captured in the dark. By using a sparse kernel module to model the blurring process, Ev-DeblurNeRF \cite{ma2022deblur} proposes a framework that could reconstruct sharp novel views of a scene using blurry images. PRDF \cite{peng2022pdrf} uses a Progressive Blur Estimation Module to obtain coarse blur estimation and then refines it using the corresponding 3D features. Despite impressive results from these prior works, they are still limited by factors such as the amount of light and motion blur in images.

\subsection{Event-based NeRFs}
Event-based NeRFs show improvement in 3D reconstruction for scenes with motion blur \cite{klenk2023nerf} \cite{qi2023e2nerf} \cite{poggi2022cross}. ENeRF \cite{klenk2023nerf} was the first attempt at tackling NeRFs using event cameras, and was able to recover sharp details from blurry frames. This work also shows some success at reconstructing scenes shot in low illumination. E$^2$NeRF \cite{qi2023e2nerf} exploits the internal relationship between events and frames, significant improving the performance of their NeRF for blurred and novel views. EvDeblurNeRF \cite{Cannici_2024_CVPR} builds on these prior works, and improves the robustness of NeRF to motion blur and sparsity of training views. They also provide an efficient multimodal NeRF architecture, incorporating past innovations. These works showcase the robustness that event cameras offer, due to their high temporal resolution and high pixel bandwidth.

\subsection{Thermal NeRFs}
Thermal NeRFs have shown impressive ability to render NeRFs in visually degraded scenarios \cite{ye2024thermal} \cite{hassan2024thermonerf}. Thermal-NeRF \cite{ye2024thermal} leverages solely thermal cameras for NeRF estimation. Using a structural thermal constraint to address challenges of sparse texture and feature scarcity in thermal images, Thermal-NeRF is able to recover detailed features. ThermoNeRF \cite{hassan2024thermonerf} combines both thermal and RGB cameras, and achieves accurate thermal view synthesis. While this work excels on thermal scene reconstruction, there is no apparent improvement for RGB scene reconstruction. X-NeRF \cite{poggi2022cross} uses RGB, thermal and multi-spectral cameras to estimate a NeRF. This work is the first to learn a cross-spectral scene representation, but struggles at cross-modal alignment. \cite{zhu2023multimodal} is a multi-modal approach showing marginal improvement for NeRFs using RGB, thermal and point cloud modalities. Thus far, multimodal NeRFs have not shown marked improvement for reconstructions across all modalities.

\section{Method}

\begin{figure*}
    \centering
\includegraphics[width=0.9\textwidth]{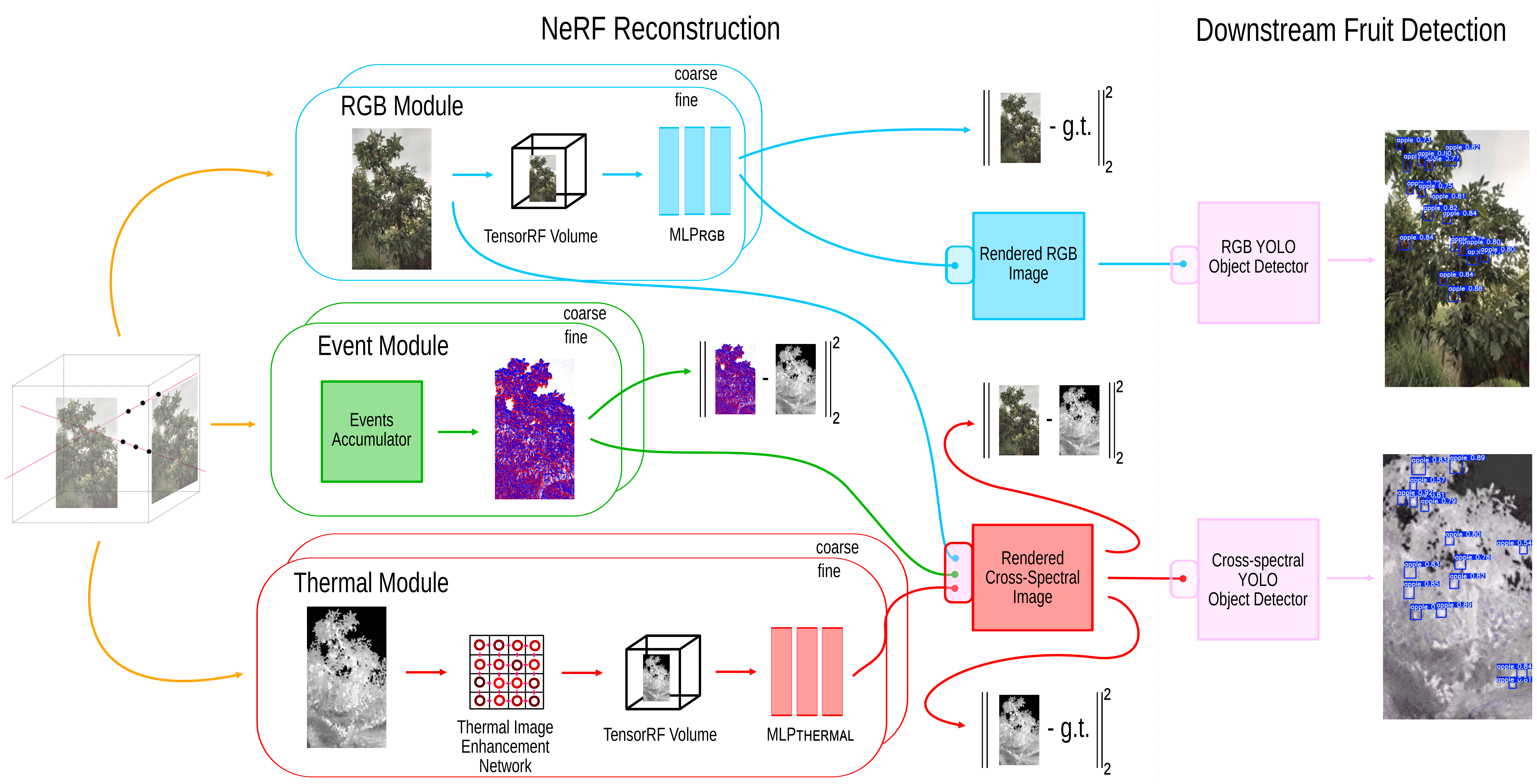}
\caption{Architecture of \textit{AgriNeRF} and downstream fruit detection. We take frames from each of the cameras and sample features from coarse and fine TensorRF volumes. These features are then passed into the $MLP$s for the RGB and thermal modules. By optimizing over a regularization loss, we render a cross-spectral reconstruction. We generate two different NeRFs which are then leveraged for robust fruit detection.}
\label{fig:figure2}
\vspace{-.5cm}
\end{figure*}

Traditional RGB-only NeRFs often struggle to accurately reconstruct scenes in challenging lighting conditions, such as low light, harsh brightness, and shadows. These limitations are especially problematic for fruit detection in agricultural settings. By leveraging a multi-modal framework that integrates RGB, thermal, and event cameras, we can create NeRFs that are robust to variations in illumination and capable of delivering more accurate reconstructions across diverse lighting scenarios. In this section we outline the methodology underlying \textit{AgriNeRF}.

\subsection{AgriNeRF}

\textit{AgriNeRF} includes three different sensing modules, incorporating data from RGB, event, and thermal cameras. We leverage a multi-modal image alignment module which projects all modalities into a unified coordinate plane. The result is a system capable of generating both RGB and cross-spectral NeRF reconstructions. The overall architecture of the system is illustrated in Fig. \ref{fig:figure2}. These reconstructions are designed to enable robust fruit detection across varying lighting conditions and at different times of the day.

\textbf{RGB Module}. Inspired by \cite{Cannici_2024_CVPR}, we sample explicit features from a TensorRF \cite{Chen2022ECCV} volume and feed this into $MLP_{RGB}$. This approach is motivated by the computational overhead incurred by incorporating three distinct modalities in \textit{AgriNeRF}. Like \cite{Cannici_2024_CVPR}, by leveraging these additional features, we can achieve faster convergence, reducing the training time. The output of this module is the color and density at each location using only RGB features. Following is the reconstruction loss for the RGB images. $\tilde{C}_{rgb}$ is the estimated color in the rendered RGB image.

\begin{equation}
L_{rgb} = \frac{1}{|\mathcal{R}|} \sum_{r \in R} \| \tilde{C}_{rgb} - C_{rgb} \|_2^2
\end{equation}

\textbf{Event Module}. We utilize a frame-based representation to process the event data. To create these frames, we accumulate events within uniformly spaced time windows aligned with the same frame rate as the RGB and thermal cameras. This approach ensures optimal time synchronization across all three sensors. By incorporating events in the cross-spectral reconstruction, we enhance feature consistency across modalities. 

\textbf{Thermal Module}. The thermal module requires additional pre-processing before the frames can be fed into the network. We pass the raw thermal frame into the thermal image enhancement network, which re-scales it to a 8-bit thermal image. This network has been adapted from Fieldscale \cite{gil2024fieldscale}, which has the shown the ability to preserve local radiometry in thermal images, as well as fine detail and spatial consistency. Such enhancements have proven beneficial for applications like object detection, making it an ideal choice for our system. After receiving the new re-scaled frame, we sample explicit features via an additional TensorRF volume and pass these to the $MLP_{THERMAL}$. Following is the reconstruction loss for the thermal frames. $\tilde{C}_{th}$ is the estimated color in the rendered thermal image.

\begin{equation}
L_{th} = \frac{1}{|\mathcal{R}|} \sum_{r \in R} \| \tilde{C}_{th} - C_{th} \|_2^2
\end{equation}

\textbf{Cross-spectral Reconstruction}. In addition to supervising the network with ground truth RGB images, we incorporate ground truth event frames and reconstructed thermal frames to enhance feature correlation across all three modalities. This collective reconstruction leverages the RGB, event, and thermal modules together.

The cross-spectral regularization loss is designed to promote feature consistency across the different modalities, leading to better object prominence. This consistency is crucial for improving downstream tasks, such as fruit detection, as can be seen in Table \ref{tab:ablation}. Following is the cross-spectral regularization loss.

\begin{equation}
L_{reg} = \frac{1}{|\mathcal{R}|} \sum_{r \in R}( \| \tilde{C}_{th} - {C}_{rgb} \|_2^2 + \| \tilde{C}_{th} - {C}_{ev} \|_2^2)
\end{equation}

\subsection{Sensors} 

\begin{figure}
    \centering
\includegraphics[width=0.48\textwidth]{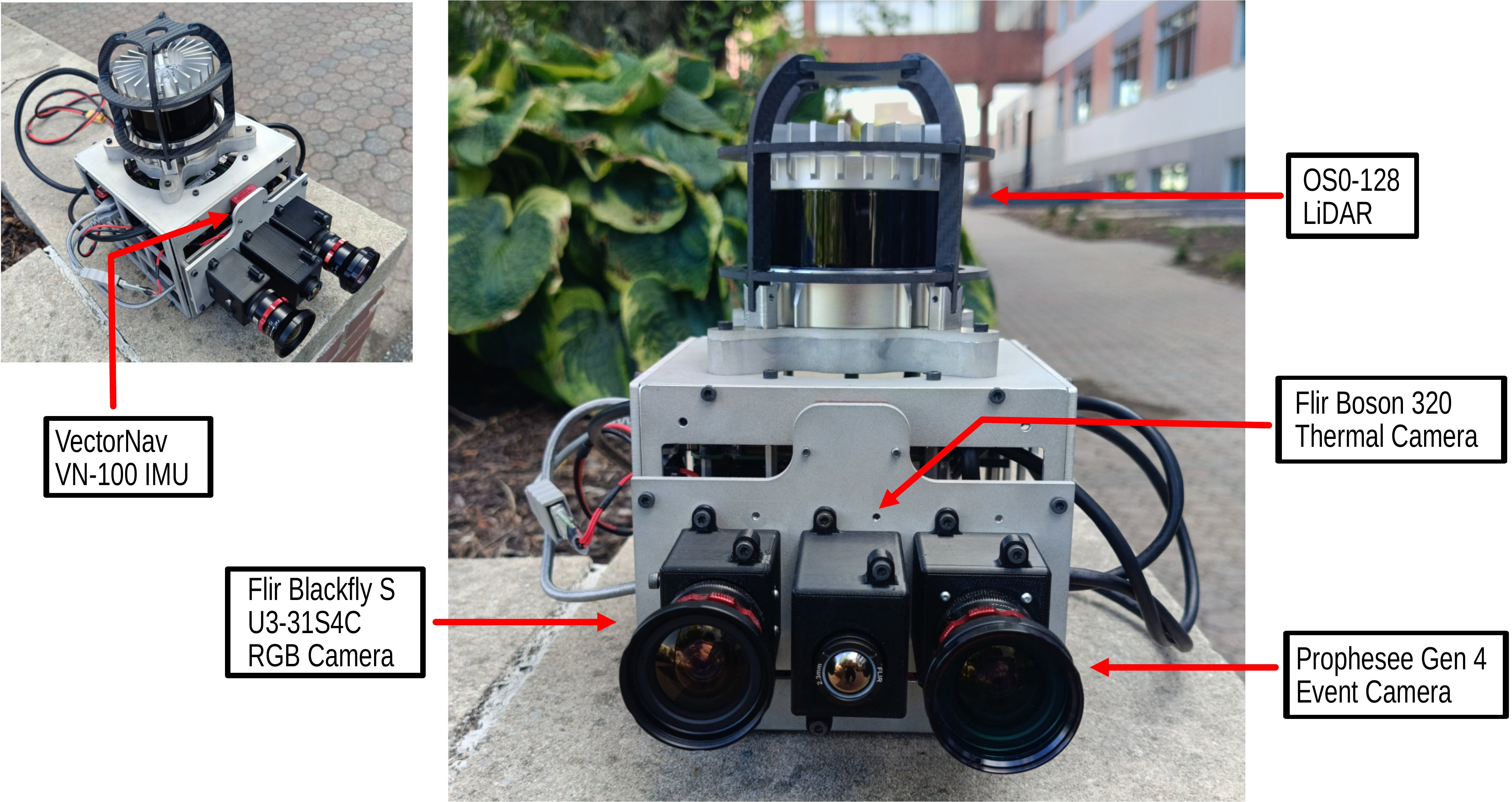}
\caption{Multimodal sensor suite incorporating RGB, thermal and event cameras used to collect datasets. We also equip the sensor suite with a LiDAR and IMU.}
\label{fig:figure3}
\vspace{-.5cm}
\end{figure}

We developed a novel multi-modal sensor suite equipped with three complementary sensors: RGB, event and thermal cameras. Specifically, our setup, in Fig. \ref{fig:figure3}, includes the FLIR Blackfly S U3-31S4C RGB camera, the Prophesee Gen 4 event camera and the FLIR Boson 320 thermal camera. These sensors were used collectively to capture the datasets for our multi-modal NeRF.

Although, our system also includes an Ouster OS0-128 LiDAR and a VectorNav VN-100 IMU, we do not use it to generate poses. Instead, we are able to use Colmap with RGB and enhanced thermal images.

\section{Experiments}

\subsection{Dataset}

We evaluate \textit{AgriNeRF} on real-world datasets collected by our multi-modal sensor suite. These datasets were gathered in diverse environments, including apple and peach orchards, as well as gardens and parks featuring tomatoes, eggplants, and peppers. These datasets were collected during various times of the day, in the morning, afternoon, evening and night, to capture different lighting conditions. 

\subsection{Implementation Details}

\textbf{NeRF reconstruction}.
We began by building our codebase upon the foundation provided by \cite{Cannici_2024_CVPR}, extending it to incorporate thermal frames and implementing regularization across the different modalities.

We use Colmap \cite{schoenberger2016sfm} to generate poses for the cameras.
In low-light scenarios, instead of relying on RGB images in COLMAP, we employed thermal images enhanced through a Thermal Image Enhancement Network. This enhancement step is based on the approach by \cite{gil2024fieldscale}, which demonstrated significant improvements in thermal image quality, making it well-suited for downstream tasks like visual place recognition and object detection.

We train \textit{AgriNeRF} for 30,000 iterations for each experiment on an Nvidia RTX A5000. We utilize a batch size of 1024 rays, and sample 64 coarse and 64 fine points along each ray. Additionally, we use 16777248 voxels for the coarse TensorRF volume,
and 134217984 voxels for the fine TensorRF volume. We use the Adam optimizer to minimize the following multi-objective loss function: $L = L_{rgb} +L_{therm} + L_{reg} $.

\textbf{Downstream fruit detection}.
For object detection, we employ YOLOv8 \cite{Jocher_Ultralytics_YOLO_2023} as well as AnyLabeling \cite{Nguyen_AnyLabeling_-_Effortless} for labelling ground truth fruits. For labelling ground truth RGB images, we simply label the original dataset images. In contrast, for the cross-spectral reconstruction, we use thermal images augmented with some cross-spectral images as the ground truth, as thermal data often serves as the dominant modality for most sequences. We train each object detection network for 250 epochs on ground truth images and then evaluate views synthesized from both the RGB and cross-spectral reconstructions.

\begin{table*}[h]
\centering
\caption{Quantitative Comparison vs Other Methods on Orchards Dataset. Best Results are reported in bold.}
\label{tab:recon_orchards} 
\begin{tabular}{|c@{\hspace{2pt}}|c@{\hspace{2pt}} c@{\hspace{2pt}} c@{\hspace{2pt}}|c@{\hspace{2pt}} c@{\hspace{2pt}} c@{\hspace{2pt}}|c@{\hspace{2pt}} c@{\hspace{2pt}} c@{\hspace{2pt}}|c@{\hspace{2pt}} c@{\hspace{2pt}} c@{\hspace{2pt}}|c@{\hspace{2pt}} c@{\hspace{2pt}} c@{\hspace{2pt}}|}
\hline
 & \multicolumn{3}{c|}{Apples Bright} & \multicolumn{3}{c|}{Peaches} & \multicolumn{3}{c|}{Apples Shadowed} & \multicolumn{3}{c|}{Average} \\
Methods & PSNR($\uparrow$) & SSIM($\uparrow$) & LPIPS($\downarrow$) & PSNR($\uparrow$) & SSIM($\uparrow$) &LPIPS($\downarrow$) & PSNR($\uparrow$) & SSIM($\uparrow$) &LPIPS($\downarrow$) & PSNR($\uparrow$) & SSIM($\uparrow$) & LPIPS($\downarrow$) \\ \hline
Nerfacto (RGB) \cite{nerfstudio} & 18.57 & 0.62 & \textbf{0.32} & 23.34 & 0.73 & \textbf{0.21} & 20.72 & 0.54 & \textbf{0.27} & 20.88 & 0.63 & \textbf{0.27} \\
\hline Nerfacto (Thermal) \cite{nerfstudio} & 13.46 & 0.36 & 0.75 & 16.80 & 0.54 & 0.60 & 16.14 & 0.62 & 0.60 & 15.47 & 0.51 & 0.65 \\
\hline
EvDeblurNeRF \cite{Cannici_2024_CVPR} & 23.01 & 0.68 & 0.52 & 26.98 & 0.75 & 0.42 & 25.54 & 0.66 & 0.43 & 25.18 & 0.70 & 0.46 \\ \hline
Ours (RGB) & \textbf{25.48} & \textbf{0.76} & 0.41 & \textbf{29.21} & \textbf{0.83} & 0.31 & \textbf{27.63} & \textbf{0.74} & 0.35 & \textbf{27.44} & \textbf{0.78} & 0.36 \\ \hline
\end{tabular}
\end{table*}

\begin{table*}[h]
\centering
\caption{Quantitative Comparison vs Other Methods on Garden Dataset. Best Results are reported in bold.}
\label{tab:recon_garden}
\begin{tabular}{|c@{\hspace{2pt}}|c@{\hspace{2pt}} c@{\hspace{2pt}} c@{\hspace{2pt}}|c@{\hspace{2pt}} c@{\hspace{2pt}} c@{\hspace{2pt}}|c@{\hspace{2pt}} c@{\hspace{2pt}} c@{\hspace{2pt}}|c@{\hspace{2pt}} c@{\hspace{2pt}} c@{\hspace{2pt}}|c@{\hspace{2pt}} c@{\hspace{2pt}} c@{\hspace{2pt}}|}
\hline
 & \multicolumn{3}{c|}{Eggplants} & \multicolumn{3}{c|}{Tomatoes} & \multicolumn{3}{c|}{Peppers} & \multicolumn{3}{c|}{Average} \\ 
Methods & PSNR($\uparrow$) & SSIM($\uparrow$) & LPIPS($\downarrow$) & PSNR($\uparrow$) & SSIM($\uparrow$) & LPIPS($\downarrow$) &PSNR($\uparrow$) & SSIM($\uparrow$) & LPIPS($\downarrow$) & PSNR($\uparrow$) & SSIM($\uparrow$) & LPIPS($\downarrow$) \\ \hline
Nerfacto (RGB) \cite{nerfstudio} & 18.62 & 0.62 & \textbf{0.30} & 24.46 & 0.86 & \textbf{0.24} & 21.64 & 0.52 & \textbf{0.44} & 21.57 & 0.67 & \textbf{0.33} \\ 
\hline
Nerfacto (Thermal) \cite{nerfstudio} & 12.44 & 0.50 & 0.72 & 18.53 & 0.56 & 0.58 & 18.29 & \textbf{0.68} & 0.46 & 16.42 & 0.58 & 0.59 \\ 
\hline
EvDeblurNeRF \cite{Cannici_2024_CVPR} & 25.29 & 0.78 & 0.50 & 31.82 & 0.91 & 0.41 & 26.38 & 0.59 & 0.66 & 27.83 & 0.76 & 0.52 \\ \hline
Ours (RGB) & \textbf{27.24} & \textbf{0.82} & 0.44 & \textbf{33.56} & \textbf{0.92} & 0.37 & \textbf{28.28} & 0.66 & 0.57 & \textbf{29.69} & \textbf{0.80} & 0.46 \\ \hline
\end{tabular}
\end{table*}

\begin{table*}[h]
\centering
\caption{Object Detection Results Against Other Methods. Best Results are reported in bold.}
\label{tab:obj_detection}
\begin{tabular}{|c@{\hspace{1.5pt}}|c@{\hspace{1.5pt}}|c@{\hspace{1.5pt}}|c@{\hspace{1.5pt}}|c@{\hspace{1.5pt}}|c@{\hspace{1.5pt}}|c@{\hspace{1.5pt}}|c|}
\hline & Methods & mAP50($\uparrow$) & mAP50-95($\uparrow$) & Precision($\uparrow$) & Recall($\uparrow$) & F1-Score($\uparrow$) & Lighting \\ \cline{1-8}
\multirow{3}{*}{\parbox{2cm} {Garden \\ Tomatoes}} & Nerfacto (RGB) \cite{nerfstudio} & 25.1 & 5.05 & 0.6 & 0.250 & 0.353 & \multirow{3}{*}{\textcolor{blue}{\faCloudMoon} Dark evening}\\ 
& Nerfacto (Thermal) \cite{nerfstudio} & 20.8 & 4.2 & 0.879 & 0.091 & 0.165 & \\ 
& Ours (RGB) & 22.9 & 11.2 & 0.286 & 0.333 & 0.308 & \\ 
& Ours (Cross-Spectral) & \textbf{75.4} & \textbf{36.9} & \textbf{0.882} & \textbf{0.577} & \textbf{0.698} & \\ \hline
\multirow{3}{*}{\parbox{2cm} {Garden \\ Eggplants}} & Nerfacto (RGB) & 64.2 & 40.3 & 0.971 & 0.429 & 0.595 & \multirow{3}{*}{\textcolor{gray}{\faMoon} Evening} \\ 
& Nerfacto (Thermal) & 66.7 & 20.0 & \textbf{1.00} & 0.333 & 0.500 & \\ 
& Ours (RGB) & 56.2 & 33.7 & \textbf{1.00} & 0.125 & 0.222 & \\ 
& Ours (Cross-Spectral) & \textbf{92.9} & \textbf{56.3} & 0.958 & \textbf{0.885} & \textbf{0.920} & \\ \hline
\multirow{3}{*}{\parbox{2cm}{Orchard Apples \\ Shadowed}} & Nerfacto (RGB) & 70.7 & 24.2 & 0.555 & \textbf{0.833} & 0.666 & \multirow{3}{*}{\textcolor{orange}{\faCloudSun} Shadowed morning} \\ 
& Nerfacto (Thermal) & 50.9 & 15.5 & 0.631 & 0.444 & 0.521 & \\
& Ours (RGB) & \textbf{80.4} & \textbf{33.8} & \textbf{0.855} & 0.718 & \textbf{0.781} & \\ 
& Ours (Cross-Spectral) & 63.3 & 29.3 & 0.614 & 0.603 & 0.608 & \\ \hline
\multirow{3}{*}{\parbox{2cm} {Average}} & Nerfacto (RGB) & 53.3 & 23.2 & 0.709 & 0.504 & 0.538\\
& Nerfacto (Thermal) & 46.1 & 13.2 & \textbf{0.834} & 0.289 & 0.395\\
& Ours (RGB) & 53.2 & 26.2 & 0.714 & 0.392 & 0.437\\
& Ours (Cross-Spectral) & \textbf{77.2} & \textbf{40.8} & 0.818 & \textbf{0.688} & \textbf{0.742}\\ \cline{1-7}
\end{tabular}
\end{table*}

\begin{table*}[h]
\centering
\caption{Object Detection Ablation on Garden Eggplants Sequence. Best Results are reported in bold.}
\label{tab:ablation}
\begin{tabular}{|c|c|c|c|c|c|c|}
\hline
Methods & mAP50 ($\uparrow$) & mAP50-95 ($\uparrow$) & Precision ($\uparrow$) & Recall ($\uparrow$) & F1-Score ($\uparrow$)\\ \hline
Ours (RGB) & 56.2 & 33.7 & \textbf{1.00} & 0.125 & 0.222\\ \hline
Ours (RGB+Events) & 84.2 & \textbf{60.7} & 0.991 & 0.706 & 0.825\\ \hline
Ours (RGB+Thermal) & 86.7 & 41.8 & \textbf{1.00} & 0.733 & 0.846\\ \hline
Ours (Thermal) & 92.3 & 44.3 & \textbf{1.00} & 0.846 & 0.917\\ \hline
Ours (Thermal+Events) & 76.2 & 41.1 & 0.818 & 0.692 & 0.749 \\ \hline
Ours (All) & \textbf{92.9} & 56.3 & 0.958 & \textbf{0.885} & \textbf{0.920}\\ \hline
\end{tabular}
\end{table*}

\subsection{Baselines}

We evaluate our scene reconstruction on three metrics: PSNR, SSIM and LPIPS. For object detection, we use the metrics of mAP50, mAP50-95, precision, recall, and F1-score.

For comparison, we benchmark our results against both EvDeblurNeRF \cite{Cannici_2024_CVPR}, and Nerfacto's \cite{nerfstudio} RGB and thermal reconstructions.

\section{Results}

In this section, we present both qualitative and quantitative results of our RGB scene reconstruction, using self-collected datasets, and compare them against other state-of-the-art methods. It is important to note that we do not evaluate our cross-spectral scene reconstruction using PSNR, LPIPS, and SSIM metrics. Instead, we assess the downstream capabilities of the cross-spectral NeRF for fruit detection, comparing its performance against other methods.

To ensure a comprehensive evaluation in various degraded lighting conditions, we use sequences captured during a dark evening, regular evening, and shadowed morning. For this analysis, we benchmark both the RGB and cross-spectral reconstructions.

Lastly, we conduct an ablation study to highlight the improved detection performance achieved by incorporating multiple modalities.

\subsection{Reconstruction Results on Orchards Dataset}

We benchmark our reconstructions both quantitatively and qualitatively, as shown in Table \ref{tab:recon_orchards} and Fig. \ref{fig:figure4}, respectively. On average, our NeRF outperforms other methods in terms of PSNR and SSIM across the majority of sequences in the Orchards datasets. Although our reconstruction ranks second to Nerfacto in LPIPS, it avoids some of the artificial artifacts present in the latter’s output. For example, in the upper half of Fig. \ref{fig:figure4}, Nerfacto’s RGB reconstruction introduces an extra cloud in the top left corner that is absent in the ground truth image. While LPIPS measures perceptual similarity, our method excels in pixel-level accuracy, resulting in a higher fidelity reconstruction.

Our RGB reconstruction achieves a notable improvement of +2.26 dB in PSNR and a +11.4\% increase in SSIM compared to the second-best method. 

\subsection{Reconstruction Results on Garden Dataset}

\begin{figure}
    \centering
\includegraphics[width=0.48\textwidth]{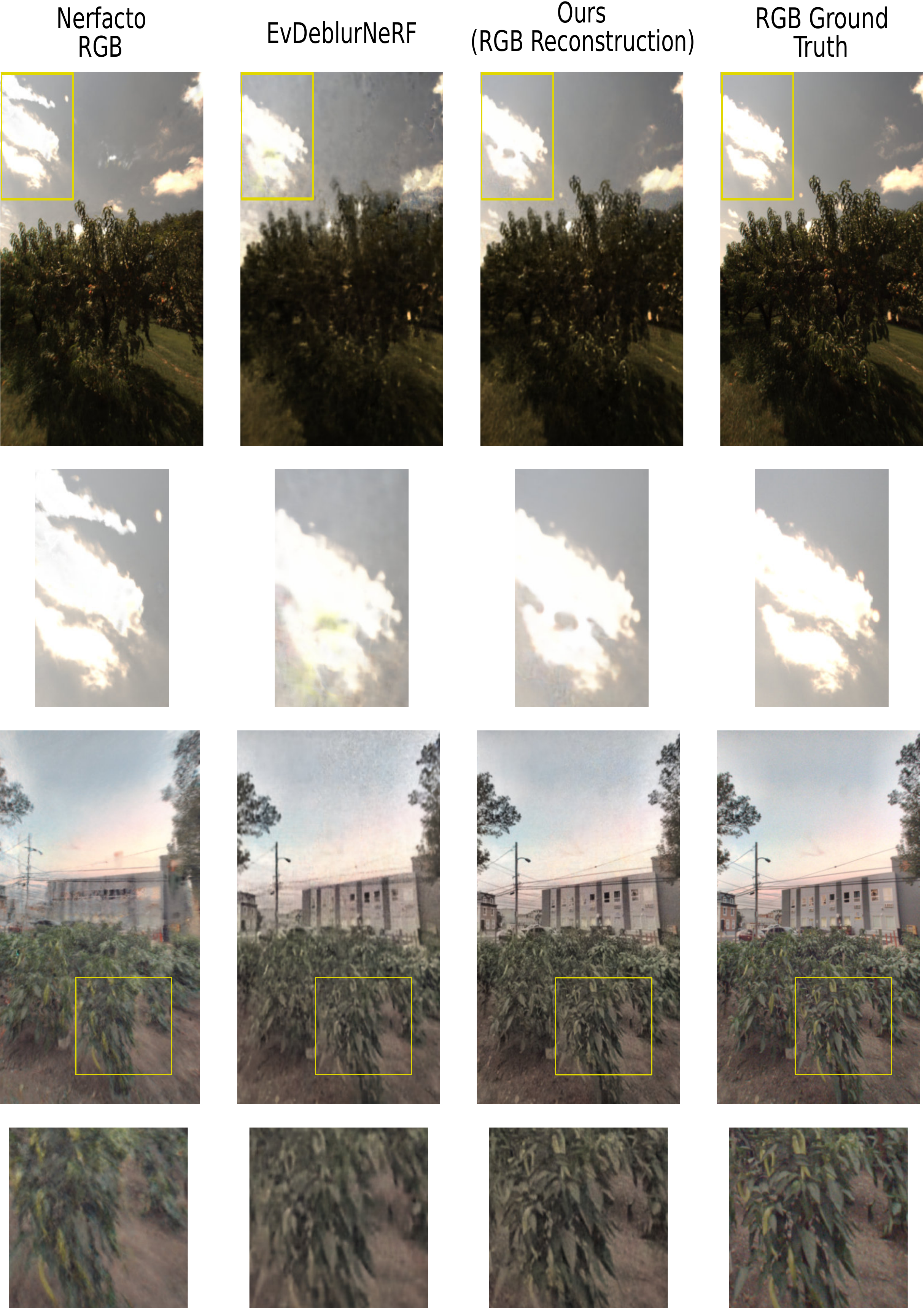}
\caption{Qualitative comparison on Orchards and Garden datasets. \underline{Top}: The Nerfacto reconstruction introduces an extra cloud, which is absent in the ground truth, while our method produces the most accurate reconstruction. \underline{Bottom}: Our approach consistently delivers the highest fidelity, effectively capturing the fine details of the peppers with superior accuracy.}
\label{fig:figure4}
\vspace{-.5cm}
\end{figure}

Similar to the results on the Orchards dataset, our method performs best on average in the PSNR and SSIM metrics on the Garden dataset, as shown in Table \ref{tab:recon_garden} and Fig. \ref{fig:figure4}. Our NeRF achieves a +1.86 dB improvement in PSNR and a +5.26\% increase in SSIM compared to the second-best method. Once again, Nerfacto outperforms all others in LPIPS, with our method coming in second. In the bottom half of Fig. \ref{fig:figure4}, our method shows a significantly better reconstruction of the scene compared to Nerfacto and EvDeblurNeRF. Our approach excels in accurately capturing the fine details of the peppers, highlighting its superior performance.

\subsection{Fruit Detection Results}

In the garden tomatoes sequence, a low-light sequence, our cross-spectral reconstruction outperforms all others, with approximately 3x better mAP50 and mAP50-95 compared to the second-best method, as shown in Table III. Similarly, for the garden eggplants sequence, another evening scene with slightly more light, our cross-spectral reconstruction achieves 1.4x higher mAP50 and mAP50-95 than the next best method. The first two rows of Fig. \ref{fig:figure5} highlight the improved fruit detection performance of the cross-spectral reconstruction over its RGB counterpart.

In contrast to the earlier low-light sequences, in the orchard apples shadowed sequence, our RGB reconstruction outperforms others in mAP50, mAP50-95, precision, and F1-score.

Overall, our cross-spectral reconstruction delivers the best fruit detection performance, with a +44.8\% improvement in mAP50, +55.7\% increase in mAP50-95, +36.5\% increase in recall, and +37.9\% increase in F1-score compared to the second-best method. The most significant gains occur in low-light sequences, while RGB reconstructions perform equally well or better in morning sequences.

\subsection{Ablation Results}

Table \ref{tab:ablation} illustrates the improvement in fruit detection metrics for the garden eggplants sequence as additional modalities are integrated into the cross-spectral reconstruction. 

\begin{figure}
    \centering\includegraphics[width=0.48\textwidth]{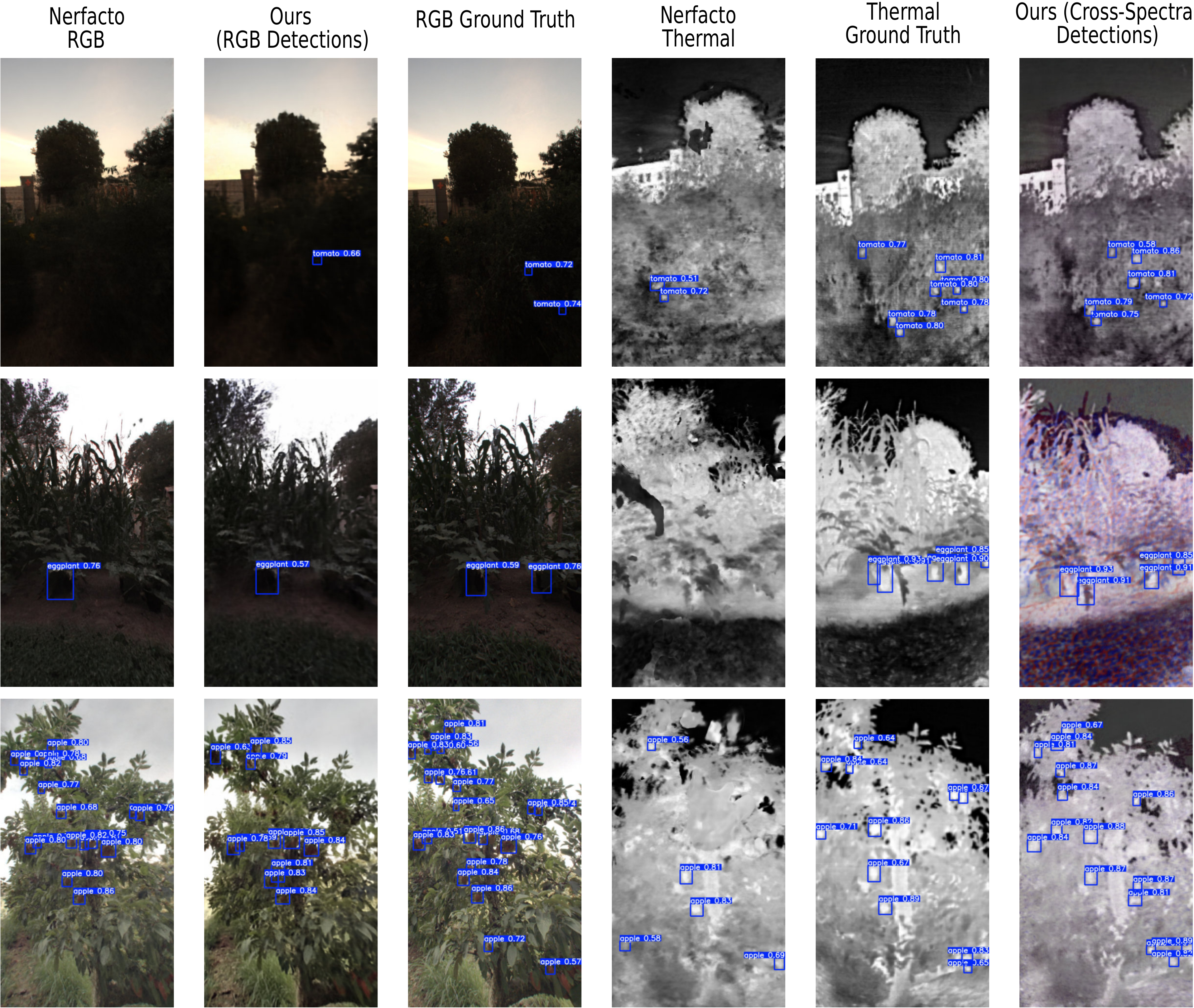}
\caption{Object detection comparison on Orchards and Garden datasets. For low-light sequences, our cross-spectral reconstruction is able to detect fruits not visible in the RGB reconstructions, thus showing highest overall performance and robustness in fruit detection.}
\label{fig:figure5}
\vspace{-.5cm}
\end{figure}

As can be seen in Fig. \ref{fig:figure5}, thermal is the dominant modality in the cross-spectral reconstruction for this sequence. In this low-light sequence, the eggplants display a distinguishable thermal signature compared to the environment, resulting in much higher performance relative to using any other single modality. Using just thermal shows only a marginal decrease in performance compared to the best method.

The best performance is achieved when all modalities—RGB, thermal, and events—are utilized in the cross-spectral reconstruction. Compared to RGB alone, we achieve a +65.3\% increase in mAP50, a +67.1\% improvement in mAP50-95, a +608\% boost in recall, and a +314\% rise in F1-score. When compared to thermal alone, our method results in a +0.65\% increase in mAP50, a +27.1\% improvement in mAP50-95, a +4.61\% increase in recall, and a +0.33\% rise in F1-score.

\section{Conclusions}

In this paper, we present AgriNeRF, a novel approach that fuses RGB, thermal, and event camera data to create both RGB and cross-spectral reconstructions for automated fruit detection.

Our multi-modal framework demonstrates substantial improvements over state-of-the-art methods, such as Nerfacto \cite{nerfstudio} and EvDeblurNeRF \cite{Cannici_2024_CVPR}. Specifically, our RGB reconstruction achieves a +2.06 dB improvement in PSNR and an +8.3\% increase in SSIM on average. Furthermore, on downstream fruit detection, our cross-spectral reconstructions shows a +44.8\% increase in mAP50, and a +55.7\% increase in mAP50-95. These enhancements highlight the increased robustness of our NeRF across diverse environments.

Our work does have limitations in night-time scenarios where there are no light or heat sources. In such environments, the RGB and event cameras are unable to capture any meaningful details. Although the thermal camera can detect coarse features, it struggles to resolve finer details, such as fruits. 
We leave it to future work to explore the use of onboard illumination as a means to resolve finer details in pitch-dark scenes.

Our cross-spectral scene reconstruction excels in fruit detection, outperforming other methods across all metrics except precision. Our ablation study validates that incorporating all modalities significantly enhances the downstream fruit detection performance of the cross-spectral reconstruction.

\addtolength{\textheight}{-9cm}   






\section*{ACKNOWLEDGMENT}

We would like to thank the research staff at Kumar Robotics, Jeremy Wang and Alex Zhou for their help in the design and manufacturing of the sensor suite. We would like to thank Jiuzhou Lei and Xu Liu, for their assistance during  dataset collection. Finally, we would like to thank the Shaw family and Wiota St. Community for allowing us to capture data in their orchards and gardens.


\bibliographystyle{IEEEtran}
\bibliography{IEEEabrv, literature}

\begin{thebibliography}{10}
\providecommand{\url}[1]{#1}
\csname url@samestyle\endcsname
\providecommand{\newblock}{\relax}
\providecommand{\bibinfo}[2]{#2}
\providecommand{\BIBentrySTDinterwordspacing}{\spaceskip=0pt\relax}
\providecommand{\BIBentryALTinterwordstretchfactor}{4}
\providecommand{\BIBentryALTinterwordspacing}{\spaceskip=\fontdimen2\font plus
\BIBentryALTinterwordstretchfactor\fontdimen3\font minus \fontdimen4\font\relax}
\providecommand{\BIBforeignlanguage}[2]{{%
\expandafter\ifx\csname l@#1\endcsname\relax
\typeout{** WARNING: IEEEtran.bst: No hyphenation pattern has been}%
\typeout{** loaded for the language `#1'. Using the pattern for}%
\typeout{** the default language instead.}%
\else
\language=\csname l@#1\endcsname
\fi
#2}}
\providecommand{\BIBdecl}{\relax}
\BIBdecl

\bibitem{tagarakis2024digital}
A.~C. Tagarakis, L.~Benos, G.~Kyriakarakos, S.~Pearson, C.~G. S{\o}rensen, and D.~Bochtis, ``Digital twins in agriculture and forestry: A review,'' \emph{Sensors}, vol.~24, no.~10, p. 3117, 2024.

\bibitem{bargoti2017deep}
S.~Bargoti and J.~Underwood, ``Deep fruit detection in orchards,'' in \emph{2017 IEEE international conference on robotics and automation (ICRA)}.\hskip 1em plus 0.5em minus 0.4em\relax IEEE, 2017, pp. 3626--3633.

\bibitem{mildenhall2021nerf}
B.~Mildenhall, P.~P. Srinivasan, M.~Tancik, J.~T. Barron, R.~Ramamoorthi, and R.~Ng, ``Nerf: Representing scenes as neural radiance fields for view synthesis,'' \emph{Communications of the ACM}, vol.~65, no.~1, pp. 99--106, 2021.

\bibitem{hu2024high}
K.~Hu, W.~Ying, Y.~Pan, H.~Kang, and C.~Chen, ``High-fidelity 3d reconstruction of plants using neural radiance fields,'' \emph{Computers and Electronics in Agriculture}, vol. 220, p. 108848, 2024.

\bibitem{fruitnerf2024}
\BIBentryALTinterwordspacing
L.~Meyer, A.~Gilson, U.~Schmidt, and M.~Stamminger, ``Fruitnerf: A unified neural radiance field based fruit counting framework,'' in \emph{IROS}, 2024. [Online]. Available: \url{https://meyerls.github.io/fruit\_nerf}
\BIBentrySTDinterwordspacing

\bibitem{gallego2020event}
G.~Gallego, T.~Delbr{\"u}ck, G.~Orchard, C.~Bartolozzi, B.~Taba, A.~Censi, S.~Leutenegger, A.~J. Davison, J.~Conradt, K.~Daniilidis \emph{et~al.}, ``Event-based vision: A survey,'' \emph{IEEE transactions on pattern analysis and machine intelligence}, vol.~44, no.~1, pp. 154--180, 2020.

\bibitem{Cannici_2024_CVPR}
M.~Cannici and D.~Scaramuzza, ``Mitigating motion blur in neural radiance fields with events and frames,'' in \emph{Proceedings of the IEEE/CVF Conference on Computer Vision and Pattern Recognition (CVPR)}, 2024.

\bibitem{klenk2023nerf}
S.~Klenk, L.~Koestler, D.~Scaramuzza, and D.~Cremers, ``E-nerf: Neural radiance fields from a moving event camera,'' \emph{IEEE Robotics and Automation Letters}, vol.~8, no.~3, pp. 1587--1594, 2023.

\bibitem{gade2014thermal}
R.~Gade and T.~B. Moeslund, ``Thermal cameras and applications: a survey,'' \emph{Machine vision and applications}, vol.~25, pp. 245--262, 2014.

\bibitem{ye2024thermal}
T.~Ye, Q.~Wu, J.~Deng, G.~Liu, L.~Liu, S.~Xia, L.~Pang, W.~Yu, and L.~Pei, ``Thermal-nerf: Neural radiance fields from an infrared camera,'' \emph{arXiv preprint arXiv:2403.10340}, 2024.

\bibitem{hassan2024thermonerf}
M.~Hassan, F.~Forest, O.~Fink, and M.~Mielle, ``Thermonerf: Multimodal neural radiance fields for thermal novel view synthesis,'' \emph{arXiv preprint arXiv:2403.12154}, 2024.

\bibitem{mildenhall2022nerf}
B.~Mildenhall, P.~Hedman, R.~Martin-Brualla, P.~P. Srinivasan, and J.~T. Barron, ``Nerf in the dark: High dynamic range view synthesis from noisy raw images,'' in \emph{Proceedings of the IEEE/CVF conference on computer vision and pattern recognition}, 2022, pp. 16\,190--16\,199.

\bibitem{ma2022deblur}
L.~Ma, X.~Li, J.~Liao, Q.~Zhang, X.~Wang, J.~Wang, and P.~V. Sander, ``Deblur-nerf: Neural radiance fields from blurry images,'' in \emph{Proceedings of the IEEE/CVF Conference on Computer Vision and Pattern Recognition}, 2022, pp. 12\,861--12\,870.

\bibitem{peng2022pdrf}
C.~Peng and R.~Chellappa, ``Pdrf: Progressively deblurring radiance field for fast and robust scene reconstruction from blurry images,'' \emph{arXiv preprint arXiv:2208.08049}, 2022.

\bibitem{lee2023dp}
D.~Lee, M.~Lee, C.~Shin, and S.~Lee, ``Dp-nerf: Deblurred neural radiance field with physical scene priors,'' in \emph{Proceedings of the IEEE/CVF Conference on Computer Vision and Pattern Recognition}, 2023, pp. 12\,386--12\,396.

\bibitem{qi2023e2nerf}
Y.~Qi, L.~Zhu, Y.~Zhang, and J.~Li, ``E2nerf: Event enhanced neural radiance fields from blurry images,'' in \emph{Proceedings of the IEEE/CVF International Conference on Computer Vision}, 2023, pp. 13\,254--13\,264.

\bibitem{poggi2022cross}
M.~Poggi, P.~Z. Ramirez, F.~Tosi, S.~Salti, S.~Mattoccia, and L.~Di~Stefano, ``Cross-spectral neural radiance fields,'' in \emph{2022 International Conference on 3D Vision (3DV)}.\hskip 1em plus 0.5em minus 0.4em\relax IEEE, 2022, pp. 606--616.

\bibitem{zhu2023multimodal}
H.~Zhu, Y.~Sun, C.~Liu, L.~Xia, J.~Luo, N.~Qiao, R.~Nevatia, and C.-H. Kuo, ``Multimodal neural radiance field,'' in \emph{2023 IEEE International Conference on Robotics and Automation (ICRA)}.\hskip 1em plus 0.5em minus 0.4em\relax IEEE, 2023, pp. 9393--9399.

\bibitem{Chen2022ECCV}
A.~Chen, Z.~Xu, A.~Geiger, J.~Yu, and H.~Su, ``Tensorf: Tensorial radiance fields,'' in \emph{European Conference on Computer Vision (ECCV)}, 2022.

\bibitem{gil2024fieldscale}
H.~Gil, M.-H. Jeon, and A.~Kim, ``Fieldscale: Locality-aware field-based adaptive rescaling for thermal infrared image,'' \emph{IEEE Robotics and Automation Letters}, 2024.

\bibitem{schoenberger2016sfm}
J.~L. Sch\"{o}nberger and J.-M. Frahm, ``{Structure-from-Motion Revisited},'' in \emph{Conference on Computer Vision and Pattern Recognition (CVPR)}, 2016.

\bibitem{Jocher_Ultralytics_YOLO_2023}
\BIBentryALTinterwordspacing
G.~Jocher, A.~Chaurasia, and J.~Qiu, ``{Ultralytics YOLO},'' Jan. 2023. [Online]. Available: \url{https://github.com/ultralytics/ultralytics}
\BIBentrySTDinterwordspacing

\bibitem{Nguyen_AnyLabeling_-_Effortless}
\BIBentryALTinterwordspacing
V.~A. Nguyen, ``{AnyLabeling - Effortless data labeling with AI support}.'' [Online]. Available: \url{https://github.com/vietanhdev/anylabeling}
\BIBentrySTDinterwordspacing

\bibitem{nerfstudio}
M.~Tancik, E.~Weber, E.~Ng, R.~Li, B.~Yi, J.~Kerr, T.~Wang, A.~Kristoffersen, J.~Austin, K.~Salahi, A.~Ahuja, D.~McAllister, and A.~Kanazawa, ``Nerfstudio: A modular framework for neural radiance field development,'' in \emph{ACM SIGGRAPH 2023 Conference Proceedings}, ser. SIGGRAPH '23, 2023.

\end{thebibliography}

\end{document}